\PassOptionsToPackage{table}{xcolor}
\pdfoutput=1
\documentclass[11pt,a4paper,xcolor=table]{article}
\usepackage[hyperref]{acl2021}
\usepackage{times}
\usepackage{latexsym}
\usepackage[table]{xcolor}
\usepackage{graphicx}

\usepackage{amsmath}
\usepackage{booktabs}
\usepackage{multirow}
\usepackage{enumitem}
\usepackage{fdsymbol}
\usepackage{float}

\usepackage{microtype}

\aclfinalcopy 

\newcommand\jsd{\textsc{pjsd}}
\newcommand\lcs{\textsc{lcs}}
\newcommand\pit{\textsc{\%-in-t}}
\newcommand\pis{\textsc{\%-in-s}}
\newcommand\jaccard{\textsc{jaccard}}
\newcommand\bleu{\textsc{bleu}}

\newcommand\gloveal{\textsc{GloVe [aligned]}}
\newcommand\gloveutt{\textsc{GloVe [utt]}}
\newcommand\bertsent{\textsc{Sentence-Bert}}
\newcommand\univ{\textsc{Universal Sentence Encoder}}

\definecolor{stringcolor}{HTML}{FFF2CC}

\title{Measuring Conversational Uptake:\\ A Case Study on Student-Teacher Interactions}

\author{{\bf Dorottya Demszky}\textsuperscript{1}\quad {\bf Jing Liu}\textsuperscript{2}\quad {\bf Zid Mancenido}\textsuperscript{3}\quad {\bf Julie Cohen}\textsuperscript{4}\\ {\bf Heather Hill}\textsuperscript{3}\quad {\bf Dan Jurafsky}\textsuperscript{1}\quad {\bf Tatsunori Hashimoto}\textsuperscript{1}\\\textsuperscript{1}Stanford University\quad \textsuperscript{2}University of Maryland \textsuperscript{3}Harvard University \textsuperscript{4}University of Virginia \\ \texttt{\{ddemszky, thashim\}@stanford.edu}}

\date{}

\begin{document}
\maketitle
\begin{abstract}
In conversation, {\em uptake} happens when a speaker builds on the contribution of their interlocutor by, for example, acknowledging, repeating or reformulating what they have said. In education, teachers’ uptake of student contributions has been linked to higher student achievement. Yet measuring and improving teachers' uptake at scale is challenging, as existing methods require expensive annotation by experts. We propose a framework for computationally measuring uptake, by (1) releasing a dataset of student-teacher exchanges extracted from US math classroom transcripts annotated for uptake by experts; (2) formalizing uptake as pointwise Jensen-Shannon Divergence (\jsd{}), estimated via next utterance classification; (3) conducting a linguistically-motivated comparison of different unsupervised measures and (4) correlating these measures with educational outcomes. We find that although repetition captures a significant part of uptake, \jsd{} outperforms repetition-based baselines, as it is capable of identifying a wider range of uptake phenomena like question answering and reformulation. We apply our uptake measure to three different educational datasets with outcome indicators. Unlike baseline measures, \jsd{} correlates significantly with instruction quality in all three, providing evidence for its generalizability and for its potential to serve as an automated professional development tool for teachers.\footnote{Code and annotated data: \url{https://github.com/ddemszky/conversational-uptake}}
\end{abstract}

\section{Introduction}
\label{sec:intro}

\begin{figure}[t!]
 \centering
   \centering
   \includegraphics[width=\linewidth]{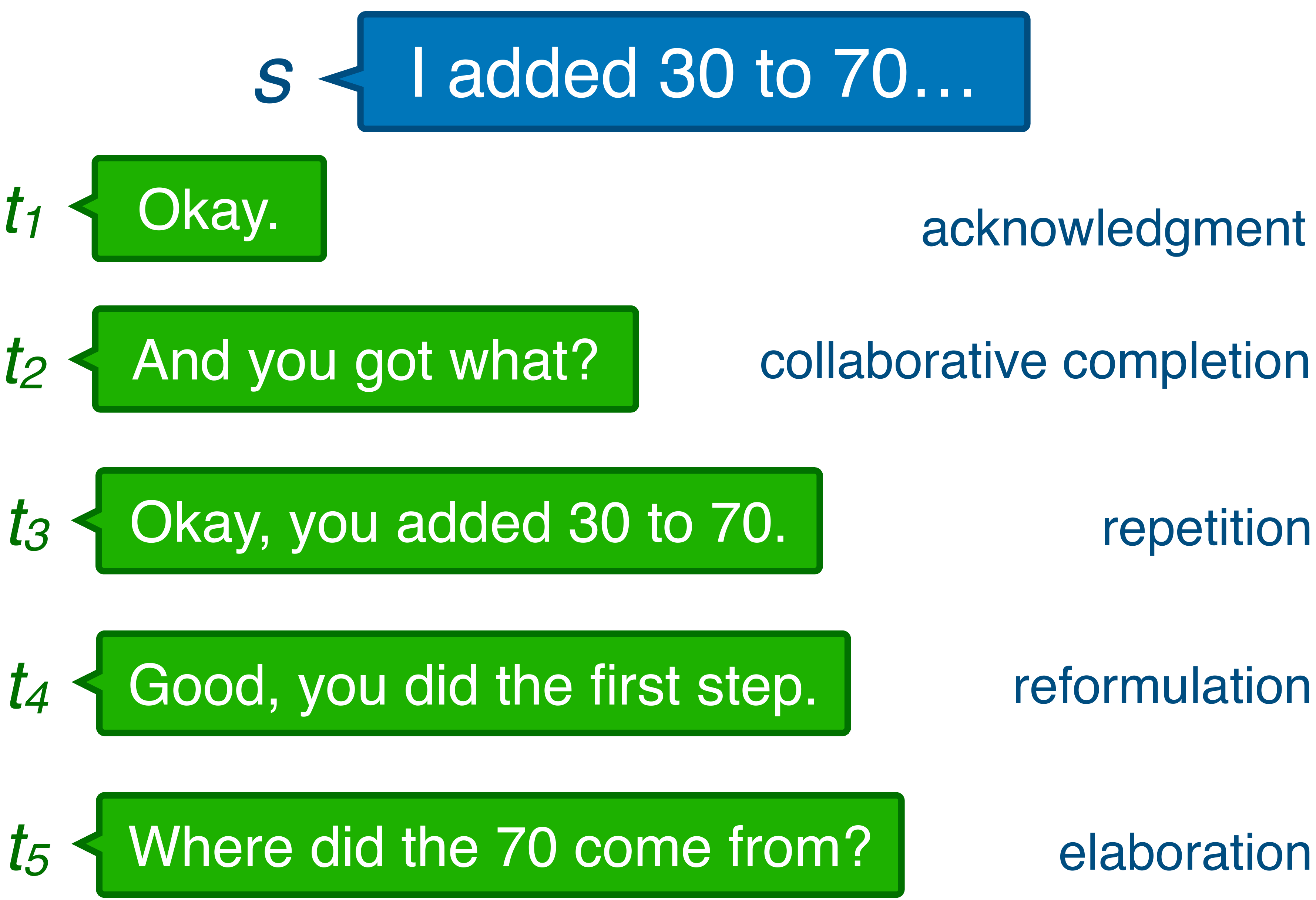}
   \caption{Example student utterance $s$ and possible teacher replies $t$, illustrating different uptake strategies.
   \label{fig:grounding_example}}
\end{figure}

Building on the interlocutor's contribution via, for example, acknowledgment, repetition or elaboration (Figure~\ref{fig:grounding_example}), is known as uptake and is key to a successful conversation. Uptake makes an interlocutor feel heard and fosters a collaborative interaction \citep{collins1982discourse,clark1989contributing}, which is especially important in contexts like education. Teachers' uptake of student ideas promotes dialogic instruction by amplifying student voices and giving them agency in the learning process, unlike monologic instruction where teachers lecture at students \citep[][]{bakhtin1986dialogic,wells1999dialogic,nystrand1997opening}. Despite extensive research showing the positive impact of uptake on student learning and achievement \citep[][]{brophy1984teacher,oconnor1993aligning,nystrand2003questions}, measuring and improving teachers' uptake at scale is challenging as existing methods require manual annotation by experts and are prohibitively resource-intensive.

We introduce a framework for computationally measuring uptake. First, we create and release a \textbf{dataset} of 2246 student-teacher exchanges extracted from US elementary math classroom transcripts, each annotated by three domain experts for teachers' uptake of student contributions.

We take an \textbf{unsupervised approach} to measure uptake in order to encourage domain-transferability and account for the fact that large amounts of labeled data are not possible in many contexts due to data privacy reasons and/or limited resources. We conduct a careful analysis of the role of \textbf{repetition} in uptake by measuring utterance overlap and similarity. We find that the proportion of student words repeated by the teacher (\pit{}) captures a large part of uptake, and that surprisingly, word-level similarity measures consistently outperform sentence-level similarity measures, including ones involving sophisticated neural models.


To capture uptake phenomena beyond repetition and in particular those relevant to teaching (e.g. question answering), we \textbf{formalize uptake} as a measure of the reply's dependence on the source utterance. We quantify dependence via \textbf{pointwise Jensen-Shannon divergence} (\jsd{}), which captures how easily someone (e.g., a student) can distinguish the true reply from randomly sampled replies. We show that \jsd{} can be estimated via cross-entropy loss obtained from next utterance classification (NUC).

We \textbf{train a model} by fine-tuning BERT-base \citep{devlin2019bert} via NUC on a large, combined dataset of student-teacher interactions and Switchboard \citep{godfrey1997switchboard}. We show that scores obtained from this model significantly outperform our baseline measures.  Using dialog act annotations on Switchboard, we demonstrate that \jsd{} is indeed better at capturing phenomena such as reformulation, question answering and collaborative completion than \pit{}, our best-performing baseline. Our manual analysis also shows qualitative differences between the models: the examples where \jsd{} outperforms \pit{} are enriched by teacher prompts for elaboration, an exemplar for dialogic instruction \citep{nystrand1997opening}.

Finally, we find that our \jsd{} measure shows a \textbf{significant linear correlation with outcomes} such as student satisfaction and instruction quality across three different datasets of student-teacher interactions: the NCTE dataset \citep{kane2015national}, a one-on-one online tutoring dataset, and the SimTeacher dataset \citep{cohen2020teacher}. These results provide evidence for the generalizability of our \jsd{} measure and for its potential to serve as an automated tool to give feedback to teachers. 

\section{Background on Uptake}
\label{sec:uptake_background}

Uptake has several linguistic and social functions. (1) It creates  \emph{coherence} between two utterances, helping structure the discourse  \citep{halliday1976cohesion,grosz1977representation,hobbs1979coherence}. (2) It is a mechanism for \emph{grounding}, i.e. demonstrating understanding of the interlocutor's contribution by accepting it as part of the common ground (shared set of beliefs among interlocutors) \citep{clark1989contributing}. (3) It promotes \emph{collaboration} with the interlocutor by sharing the floor with them and indicating what they have said is important \citep{bakhtin1986dialogic,nystrand1997opening}.

There are multiple linguistic strategies for uptake, such as acknowledgment, collaborative completion, repetition, and question answering --- see Figure~\ref{fig:grounding_example} for a non-exhaustive list. A speaker can use multiple strategies at the same time, for example, $t_3$ in Figure~\ref{fig:grounding_example} includes both acknowledgment and repetition. Different strategies can represent lower or higher uptake depending on how effectively they achieve the aforementioned functions of uptake. For example, \citet{tannen1987repetition} argues that repetition is a highly pervasive and effective strategy for ratifying listenership and building a coherent discourse. In education, high uptake has been defined as cases where the teacher follows up on the student's contribution via a question or elaboration \citep{collins1982discourse,nystrand1997opening}.

We build on this literature from discourse analysis and education to build our dataset, to develop our uptake measure and to compare the ability of different measures to capture key uptake strategies.

\section{A New Educational Uptake Dataset}
\label{sec:data}

Despite the substantial literature on the functions of uptake, we are not aware of a publicly available dataset labeled for this phenomenon. To address this, we recruit domain experts (math teachers and raters trained in classroom observation) to annotate a dataset of exchanges between students and teachers. The exchanges are sampled from transcripts of 45-60 minute long 4th and 5th grade elementary math classroom observations collected by the National Center for Teacher Effectiveness (NCTE) between 2010-2013 \citep{kane2015national}. The transcripts represent data from 317 teachers across 4 school districts in New England that serve largely low-income, historically marginalized students. Transcripts are fully anonymized: student and teacher names are replaced with terms like ``Student'', ``Teacher'' or ``Mrs. H''.\footnote{Parents and teachers gave consent for the study (Harvard IRB \#17768), and for de-identified data to be retained and used in future research. The transcripts were anonymized at the time they were created.}

\paragraph{Preparing utterance pairs.} We prepare a dataset of utterance pairs $(S,T)$, where $S$ is a student utterance and $T$ is a subsequent teacher utterance. The concept of uptake presupposes that there is something to be taken up; in our case that the student utterance has substance. For example, short student utterances like ``yes'' or ``one-third'' do not present many opportunities for uptake. Based on our pilot annotations, these utterances are difficult for even expert annotators to label. Therefore, we only keep utterance pairs where $S$ contains at least 5 tokens, excluding punctuation. We also remove all utterance pairs where the utterances contain an [Inaudible] marker, indicating low audio quality. Out of the remaining 55k $(S, T)$ pairs, we sample 2246 for annotation.\footnote{To enable potential analyses on the temporal dynamics of uptake, we randomly sampled 15 transcripts where we annotate all $(S, T)$ pairs (constituting 29\% of our annotations). The rest of the pairs are sampled from the remaining data.}

\paragraph{Annotation.} Given that uptake is a subjective and heterogeneous construct, we relied heavily on domain-expertise and took several other quality assurance steps for the annotation. As a result, the annotation took six months to develop and complete, longer than most other annotations in NLP for a similar data size ($\sim$2k examples).

Our annotation framework for uptake is designed by experts in math quality instruction, including our collaborators, math teachers and raters for the Mathematical Quality Instruction (MQI) coding instrument, used to assess math instruction \citep{learning2011measuring}. In the annotation interface, raters can see (1) the utterance pair $(S,T)$, (2) the lesson topic, which is manually labeled as part of the original dataset, and (3) two utterances immediately preceding $(S,T)$ for context. Annotators are asked to first check whether $(S,T)$ relates to math -- e.g. ``Can I go to the bathroom?'' is unrelated to math. If both $S$ and $T$ relate to math, raters are asked to select among three labels: ``low'', ``mid'' and ``high'', indicating the degree to which a teacher demonstrates that they are following what the student is saying or trying to say. The annotation framework is included in Appendix~\ref{sec:appendix_annotation}.

We recruited expert raters (with experience in teaching and classroom observation) whose demographics were representative of US K-12 teacher population. We followed standard practices in education for rater training and calibration. We conducted several pilot annotation rounds (5+ rounds with a subset of raters, 2 rounds involving all 13 raters), quizzes for raters, thorough documentation with examples, and meetings with all raters. After training raters, we randomly assign each example to three raters.

\paragraph{Post-processing and rater agreement.} Table~\ref{tab:annotated_examples} includes a sample of our annotated data. Inter-rater agreement for uptake is Spearman $\rho=.474$ (Fleiss $\kappa=.286$\footnote{We prefer to use correlations because kappa has undesirable properties \citep[see][]{delgado2019cohen} and correlations are more interpretable and directly comparable to our models' results (see later sections).}), measured by (1) excluding examples where at least one rater indicated that the utterance pair does not relate to math\footnote{This step is motivated by widely used education observation protocols such as MQI, which also clearly separate on- vs off-task instruction.}; (2) converting rater's scores into numbers (``low'': 0, ``mid'': 1, ``high'': 2); (3) z-scoring each rater's scores; (4) computing a leave-out Spearman $\rho$ for each rater by correlating their judgments with the average judgments of the other two raters; and (5) taking the average of the leave-out correlations across raters. Our interrater agreement values comparable to those obtained in widely-used classroom observation protocols such as MQI and the Classroom Assessment Scoring System (CLASS) \citep{pianta2008classroom} that include parallel measures to our uptake construct (see \citet{kelly2020using} for a summary).\footnote{High interrater variability --- especially when it comes to ratings of teacher quality --- are widely documented by gold standard studies in the field of education (see \citet{cohen2016building} for a summary).
} We obtain a single label for each example by averaging the z-scored judgments across raters.

\begin{table}
\centering
\resizebox{\linewidth}{!}{
\begin{tabular}{@{}ll@{}}
\toprule
\textbf{Example}                                                                                                                                                                                   & \textbf{Uptake}              \\ \midrule
\begin{tabular}[c]{@{}l@{}}S: ’Cause you took away 10 and 70 minus 10 is 60.\\ T:  Why did we take away 10?\end{tabular}                                                                  & {\color[HTML]{009901} high} \\ \midrule
\begin{tabular}[c]{@{}l@{}}S: There’s not enough seeds.\\ T: There’s not enough seeds.  How do you know\\right away that 128 or 132 or whatever\\it was you got doesn’t make sense?  \end{tabular}                                                                  & {\color[HTML]{009901} high} \\ \midrule
\begin{tabular}[c]{@{}l@{}}S:  Teacher L, can you change your dimensions\\like 3-D and stuff for your bars? \\ T: You can do 2-D or 3-D, yes.  I already said that.\end{tabular}                                             & {\color[HTML]{F56B00} mid}  \\ \midrule
\begin{tabular}[c]{@{}l@{}}S:  The higher the number, the smaller it is.\\ T: You got it.  That’s a good thought.\end{tabular}                                             & {\color[HTML]{F56B00} mid}  \\ \midrule
\begin{tabular}[c]{@{}l@{}}S: An obtuse angle is more than 90 degrees.\\ T:  Why don’t we put our pencils down and just do \\ some brainstorming, and then we’ll go back\\through it?\end{tabular} & {\color[HTML]{CB0000} low}  \\  \midrule
\begin{tabular}[c]{@{}l@{}}S: Because the base of it is a hexagon.\\ T: Student K?\end{tabular} & {\color[HTML]{CB0000} low}  \\ \bottomrule
\end{tabular}
}

\caption{Examples from our annotated data, showing the majority label for each example.\label{tab:annotated_examples}}
\end{table}
\section{Uptake as Overlap \& Similarity}

As we see in Table~\ref{tab:annotated_examples}, examples labeled for high uptake tend to have overlap between $S$ and $T$; this is expected, since incorporating the previous utterance in some form is known to be an important aspect of uptake (Section~\ref{sec:uptake_background}). Therefore, we begin by carefully analyzing repetition and defer discussion of more complex uptake phenomena to Section~\ref{sec:jsd}.

To accurately quantify repetition-based uptake, we evaluate a range of metrics and surprisingly find that \emph{word overlap} based measures correlate significantly better with uptake annotations than more sophisticated, utterance-level similarity measures.\footnote{We focus on unsupervised methods that enable scalability and domain-generalizability; please see Appendix~\ref{sec:appendix_supervised} for supervised baselines.}

\subsection{Methods}

We use several algorithms to better understand if word- or utterance-level similarity is a better measure of uptake. For each token-based algorithm, we experiment with several different choices for pre-processing as a way to get the best possible baselines to compare to. We include symbols for the set of choices yielding best performance
:  removing punctuation $\spadesuit$, removing stopwords using NLTK \citep{bird-2006-nltk} $\oplus$, and stemming via NLTK's SnowballStemmer $\dagger$.

\paragraph{String- and token-overlap.}

\begin{description}
[itemindent=2pt,leftmargin=3pt,parsep=0pt]

\item[\lcs:] Longest Common Subsequence.


\item[\pit:] Fraction of tokens from $S$ that are also in $T$ \citep{MillerBeebe56}. $[\spadesuit \oplus \dagger]$

\item[\pis:] Fraction of tokens from $T$ that are also in $S$. $[\spadesuit \oplus]$

\item[\jaccard:] Jaccard similarity \citep{niwattanakul2013using}. $[\spadesuit \oplus]$

\item[\bleu:] BLEU score \citep{papineni2002bleu} for up to 4-grams. We use $S$ as the reference and $T$ as the hypothesis.$[\spadesuit \oplus \dagger]$

\end{description}

\paragraph{Embedding-based similarity.}
For the word vector-based metrics, we use 300-dimensional GloVe vectors \citep{pennington2014glove} pretrained on 6B tokens from Wikipedia 2014 and the Gigaword 5 corpus \citep{parker2011english}.

\begin{description}
[itemindent=2pt,leftmargin=3pt,parsep=0pt]

\item[\gloveal:] Average pairwise cosine similarity of word embeddings between tokens from $S$ and its most similar token in $T$. $[\spadesuit]$

\item[\gloveutt:] Cosine similarity of utterance vectors representing $S$ and $T$. Utterance vectors are obtained by averaging word vectors from $S$ and from $T$. $[\spadesuit \oplus]$

\item[\bertsent:] Cosine similarity of utterance vectors representing $S$ and $T$, obtained using a pre-trained Sentence-BERT model for English \citep{reimers2019sentence}.\footnote{\url{https://github.com/UKPLab/sentence-transformers}}

\item[\univ:] Inner product of utterance vectors representing $S$ and $T$, obtained using a pre-trained Universal Sentence Encoder for English \citep{cer2018universal}.

\end{description}

\subsection{Results}

\begin{table}[]
\centering
\resizebox{\linewidth}{!}{
\begin{tabular}{@{}lll@{}}
\toprule
\textbf{Model}                                     & \multicolumn{1}{c}{$\boldsymbol{\rho}$} & \multicolumn{1}{c}{\textbf{95\% CI}} \\ \midrule
\lcs                        & .283                                    & {[}.240, .329{]}                  \\
\pit               & \textbf{.523}***                                    & {[}.488, .559{]}                  \\
\pis              & .440                                    & {[}.399, .480{]}                  \\

\jaccard                    & .450                                     & {[}.413, .487{]}                  \\
\bleu                       & .510                                    & {[}.472, .543{]}                  \\  \midrule
\gloveal        & .518                                   & {[}.483, .550{]}                  \\
\gloveutt            & .424                                    & {[}.378, .465{]}                  \\
\bertsent               & .390                                    & {[}.350, .432{]}                  \\
\univ & .448                                    & {[}.408, .486{]}                  \\ \bottomrule

\end{tabular}
}

\caption{Results from our baseline measures. Asterisks indicate that \pit{} significantly outperforms \gloveal{} ($p<0.001$), measured by a paired bootstrap test, comparing the difference between the $\rho$ obtained by \pit{} and the one by \gloveal{} across 1000 iterations, then using a t-test.
\label{tab:overlap_similarity_results}}
\end{table}
 We compute correlations between model scores and human labels via Spearman rank order correlation $\rho$. We perform bootstrap sampling (for 1000 iterations) to compute 95\% confidence intervals.

 The results are shown in Table~\ref{tab:overlap_similarity_results}. Overall, we find that token-based measures outperform utterance-based measures, with \pit{} ($\rho=.523$), \gloveal{} ($\rho=.518$) (a soft word overlap measure) and \bleu{} ($\rho=.510$) performing the best.
 Even embedding-based algorithms that are computed at the utterance-level do not outperform \pit{}, a simple word overlap baseline. It is noteworthy that all measures have a significant correlation with human judgments.

The surprisingly strong performance of \pit{}, \gloveal{} and \bleu{} provide further evidence that the extent to which $T$ repeats words from $S$ is important for uptake \citep{tannen1987repetition}, especially in the context of teaching. The fact that removing stopwords helps these measures suggests that the repetition of function words is less important for uptake; an interesting contrast to linguistic style coordination in which function words play a key role \citep{danescu2011chameleons}. Moreover, the amount of words $T$ adds in addition to words from $S$ also seems relatively irrelevant based on the lower performance of the measures that penalize $T$ containing words that are not in $S$ --- examples in Table~\ref{tab:annotated_examples} also support this result.
\section{Uptake as Dependence}
\label{sec:jsd}


Now we introduce our main uptake measure, used to capture a broader range of uptake phenomena beyond repetition including, e.g., acknowledgment and question answering (Section~\ref{sec:uptake_background}). We formalize uptake as dependence of $T$ on $S$, captured by the Jensen-Shannon Divergence, which quantifies the extent to which we can tell whether $T$ is a response to $S$ or is it a random response ($T'$). If we cannot tell the difference between $T$ and $T'$, we argue that there can be no uptake, as $T$ fails all three functions of coherence, grounding and collaboration.



We can formally define the dependence for a single teacher-student utterance pair $(s,t)$ in terms of a pointwise variant of JSD (\jsd{}) as
\newcommand{\EE}{\mathbb{E}}
\newcommand{\pjsd}{\operatorname{pJSD}}
\newcommand{\prob}{\operatorname{P}}
\newcommand{\loss}{\operatorname{L}}
\newcommand\myeq{\mkern1.5mu{=}\mkern1.5mu}
\begin{multline}
\small
    \pjsd(t,s) :=  -\frac{1}{2} \bigg(\log \prob(Z\myeq1 | M\myeq t, s) \\
    \hspace*{-.2cm} +\EE \log (1-\prob(Z\myeq1 | M\myeq T', s))\bigg) + \log(2)
    \label{eq:jsd}
\end{multline}
%
where $(S,T)$ is a teacher-student utterance pair, $T'$ is a randomly sampled teacher utterance that is independent of $S$, and  $M := ZT+(1-Z)T'$ is a mixture of the two with a binary indicator variable $Z \sim \text{Bern}(p\myeq0.5)$.

This pointwise measure relates to the standard JSD for $T|S\myeq s$ and $T'$ by taking expectations over the teacher utterance via $\EE[\mathrm{pJSD}(T,s)|S\myeq s] \myeq \mathrm{JSD}(T|S\myeq s \| T')$. We consider the pointwise variant for the rest of the section, as we are interested in a measure of dependence between a specific $(t,s)$ rather than one that is averaged over multiple teacher utterances.

\subsection{Next Utterance Classification}
\label{ssec:nuc}
The definition of \jsd{} naturally suggests an estimator based on the \emph{next utterance classification} task --- a task previously used in neighboring NLP areas like dialogue generation and   discourse coherence. We fine-tune a pre-trained BERT-base model \citep{devlin2019bert} on a dataset of $(S,T)$ pairs to predict if a specific $(s,t)$ is a true pair or not (i.e., whether $t$ came from $T$ or $T'$). The objective function is cross-entropy loss, computed over the output of the final classification layer that takes in the last hidden state of $t$. Let $Z$ be a binary indicator variable representing the model's prediction. Then, the cross entropy loss for identifying $z$ is
\begin{equation}
\loss(t,s) = -\log f_\theta(t,s) - \EE \log (1-f_\theta(T', s))
\end{equation}
Which can be used directly as an estimator for the log-probability terms in Equation~\ref{eq:jsd},
\begin{equation}
    \widehat{\pjsd}(t,s) := \dfrac{1}{2} \loss(t,s) + \log2.
\end{equation}
Standard variational arguments \cite{nowozin2016f} show that any classifier $f_\theta$ forms a lower bound on the JSD,
\[\operatorname{JSD}(T|S=s\|T') \geq \EE[\widehat{\pjsd}(T,s)|S=s].\]
Thus, our overall procedure is to fit $f_\theta(t,s)$ by maximizing $\EE[\widehat{\pjsd}(t,s)]$ over our dataset and then use $f_\theta(t,s)$ (a monotone function of $\widehat{\pjsd}(t,s)$) as our pointwise measure of dependence. 

\paragraph{Training data.} We use $(S, T)$ pairs from three sources to form our training data: the NCTE dataset \cite{kane2015national} (Section~\ref{sec:data}), Switchboard \citep{godfrey1997switchboard} and a one-on-one online tutoring dataset (Section~\ref{sec:outcomes}) --- we use a combination of datasets instead of one dataset in order to support the generalizability of the model. Filtering out examples with $S < 5$ tokens or [Inaudible] markers (Section~\ref{sec:data}), our resulting dataset consists of 259k $(S, T)$ pairs. For each $(s, t)$ pair, we randomly select 3 negative $(s, t')$ pairs from the same source dataset, yielding 777k examples.\footnote{We do not split the data into training and validation sets, as we found that using predictions on the training data vs those on the test data as our uptake measure yield similar results, so we opted for maximizing training data size.}

\paragraph{Parameter settings.} We fine-tune our model for 1 epoch to avoid overfitting with a batch size of 32 $\times$ 2 gradient accumulation steps, max length of 120 tokens for $S$ and $T$ each (the rest is truncated), learning rate of 6.24e-5 with linear decay and the AdamW optimizer \citep{loshchilov2017decoupled}. Training took about 13hrs on a single TitanX GPU.

\subsection{Results \& Analysis}
\label{ssec:nuc_results}

\begin{table}[]
\centering
\resizebox{.65\linewidth}{!}{
\begin{tabular}{@{}lll@{}}
\toprule
\textbf{Model}                                     & \multicolumn{1}{c}{$\boldsymbol{\rho}$} & \multicolumn{1}{c}{\textbf{95\% CI}} \\ \midrule

\pit               & .523                                    & {[}.488, .559{]}                  \\

\jsd & \textbf{.540}***                                    & {[}.505, .574{]}                  \\ \bottomrule
\end{tabular}
}

\caption{Results from the \jsd{} model. The asterisks, calculated as in Table~\ref{tab:overlap_similarity_results}, indicate that the difference between the two models' performance is significant. \label{tab:nuc_results}}
\end{table}

\begin{table*}
\resizebox{\linewidth}{!}{%
\begin{tabular}{@{}clc|cc}
\toprule
 & \multirow{2}{*}{\textbf{Example}}                                                                                                                                                                                                                                                                                                                                                       & \multirow{2}{*}{\begin{tabular}[c]{@{}c@{}}\textbf{Label}\\(quartile)\end{tabular}}  & \multicolumn{2}{c}{\textbf{Model predictions}} \\
                                      &                                                                                                                                                                       &                                                                                                                                                                                                                     & \jsd{}                   & \pit{}                   \\ \midrule
    1     & \begin{tabular}[c]{@{}l@{}}S: i knew that eight was a composite number and -\\ T: why? how? how did you know it was composite?\end{tabular}      & \multirow{10}{*}{top}                                                                & \cellcolor[HTML]{57C4AD}top                    & \cellcolor[HTML]{F3F3F3}mid                   \\[12pt]
                          2            & \begin{tabular}[c]{@{}l@{}}S: do you have to know division to do fractions?\\ T: i would think - division, sometimes, yes, you do need to know division to do some \\types of fractions. when we get to putting your fraction in simplest forms, yes, you\\need to know division and multiplication facts. you know something else you can find\\that comes in fractions?\end{tabular} & & \cellcolor[HTML]{57C4AD}top                    & \cellcolor[HTML]{57C4AD}top                   \\[30pt]
                          3            & \begin{tabular}[c]{@{}l@{}}S: you put a one instead of a two.\\
T: yes i did. thank you. you always correct me. that's too high. let's bring it down.\\how many times do you think, student d?\end{tabular} & & \cellcolor[HTML]{EDA247}bottom                    & \cellcolor[HTML]{EDA247}bottom                   \\\midrule
4      & \begin{tabular}[c]{@{}l@{}}S: five, six, seven, eight, you take eight off.\\ T: no, no, no equal pieces. right? okay so how many equal pieces do you need to make?\end{tabular}                                                                                                                                                                                                  &  \multirow{7}{*}{bottom}      & \cellcolor[HTML]{57C4AD}bottom                 & \cellcolor[HTML]{57C4AD}bottom                \\[15pt]
5 & \begin{tabular}[c]{@{}l@{}}S: i can prove it that it's three hundred.\\ T: and you think it's -?\end{tabular}                                                                                                                            &                                                                                                                                                  & \cellcolor[HTML]{F3F3F3}mid                    & \cellcolor[HTML]{57C4AD}bottom                \\[12pt]
                6                      & \begin{tabular}[c]{@{}l@{}}S: oh, i see it. i see it.\\ T: okay, now this is also another equivalent fraction. after you color, see if you see the\\equivalent fraction. let's see what you've got, student y.\end{tabular}                                                                                                                  &                                           & \cellcolor[HTML]{F3F3F3}mid                    & \cellcolor[HTML]{EDA247}top                  \\ \bottomrule
\end{tabular}
}
\caption{Example model predictions, comparing the \jsd{} model to \pit{}. All labels are converted to percentiles: top (75th), mid (25-75th) and bottom (25th).  Green indicates correct predictions, red indicates predictions from the opposite quartile and grey indicates mid-range predictions. \label{tab:error_analysis}}
    
\end{table*}

Table~\ref{tab:nuc_results} shows that the \jsd{} model ($\rho=.540$) significantly outperforms \pit{}. Our rough estimate on the upper bound of rater agreement ($\rho=.539$, obtained from a pilot annotation where all 13 raters rated 70 examples) indicate that our best models' scores in a similar range as human agreement.\footnote{Human agreement and model scores are not directly comparable. The human agreement values (as reported here for 13 raters and in Section~\ref{sec:data} for 3 raters) are averaged leave-out estimates across raters (skewed downward). The models’ scores represent correlations with an averaged human score, which smooths over the interrater variance of 3 raters.}

 Table~\ref{tab:error_analysis} includes illustrative examples for model predictions. Our qualitative comparison of \jsd{} and \pit{} indicates that (1) the capability of \jsd{} to differentiate between more and less important words in terms of uptake (Examples 1 and 6) accounts for many cases where \jsd{} is more accurate than \pit{}, (2) neither model is able to capture rare and semantically deep forms of uptake (Example 3), (3) \jsd{} generally gives higher scores than \pit{} to coherent responses with limited word overlap (Example 5).
 
 Now we turn to our motivating goals for proposing \jsd{} and quantitatively analyze its ability to capture more sophisticated forms for uptake.

\begin{table}[t!]
\centering
\resizebox{\linewidth}{!}{
\begin{tabular}{@{}ll@{}}
\toprule
\textbf{Label}                                                         & \textbf{Examples}                                                                                                                                                                                                                                                                                                                 \\ \midrule
\begin{tabular}[c]{@{}l@{}}elaboration\\ prompt\\ (4.25*)\end{tabular} & \begin{tabular}[c]{@{}l@{}}S: so it means that the whole equation\\ is only the same.\\ T: what does it mean? i still don't\\ understand what is it?\end{tabular}                                                                                                                                                                 \\ \midrule
\begin{tabular}[c]{@{}l@{}}reformulation\\ (2.6)\end{tabular}          & \begin{tabular}[c]{@{}l@{}}S: multiplication is like, say, for instance,\\ nine times twenty. you just take - nine just\\ nine times and add it up.\\ T: okay, so repeated addition.\\\end{tabular} \\ \midrule
\begin{tabular}[c]{@{}l@{}}answer\\ (2.67)\end{tabular}                & \begin{tabular}[c]{@{}l@{}}S: do we look at the d or the m first?\\ T: the m. what's this called, that i'm writing?\end{tabular}                                                                                                                                                                                                  \\ \midrule
\begin{tabular}[c]{@{}l@{}}collaborative\\completion (0)\end{tabular}               & \begin{tabular}[c]{@{}l@{}}S: we had to add twenty-four plus twenty-four.\\ T: because there are how many triangles?\end{tabular}                                                                                                                                                                                                 \\ \bottomrule
\end{tabular}
}

\caption{Examples for linguistic phenomena, manually labeled in the dataset where \jsd{} and \pit{} make significantly different predictions. Parenthetical numbers after the labels represent the odds ratio of examples with this label occurring in the set where \jsd{} performs better over the set where \pit{} performs better (*: $p < 0.05$, computed via a Fisher exact test). \label{tab:ncte_linguistic_phenomena}}
\end{table}

\paragraph{Comparison of linguistic phenomena.} To understand if there is a pattern explaining \jsd{}'s better performance, we quantify the occurence of different linguistic phenomena for examples where \jsd{} outperforms \pit{}. Concretely, we compute the residuals for each model, regressing the human labels on their predictions. Then, we take those examples where the difference between the two models' residuals is 1.5 standard deviations above the mean difference between their residuals. We label teacher utterances in these examples for four linguistic phenomena associated with uptake and good teaching (elaboration prompt, reformulation, collaborative completion, and answer to question), allowing multiple labels (e.g. elaboration prompt and completion often co-occur).\footnote{We label examples with above average uptake scores, as there is no trivial interpretation for uptake strategies labeled on low-uptake examples.} As Table~\ref{tab:ncte_linguistic_phenomena} shows, elaboration prompts, which are exemplars of high uptake in teaching \citep{nystrand1997opening} are significantly more likely to occur in this set --- suggesting that there is a qualitative difference between what these models capture that is relevant for teaching. We do not find a significant difference in the occurrence of reformulations, collaborative completions and answers between the two sets, possibly due to the small sample size (n=67). To see whether these differences are significant on a larger dataset, we now turn to the Switchboard dialogue corpus.

\begin{figure}[]
 \centering
   \centering
  \includegraphics[width=\linewidth]{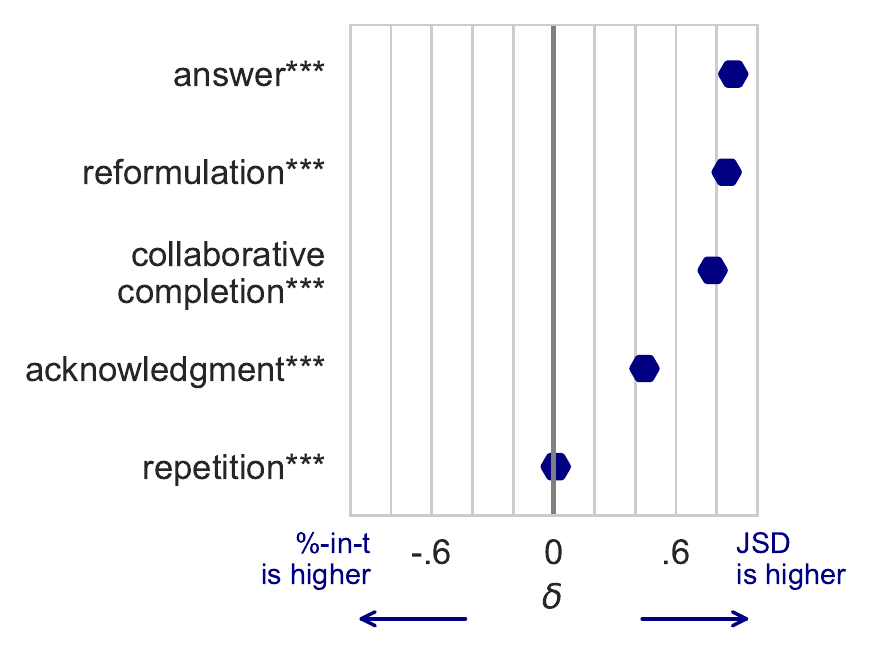}
   \caption{The difference ($\delta$) between the scores from \pit{} and \jsd{} for five uptake phenomena labeled in Switchboard. Asterisks indicate significance (***: $p < 0.001$), estimated via a median test.}
   \label{fig:swbd_damsl}
\end{figure}

\paragraph{Switchboard dialog acts.} We take advantage of dialog act annotations on Switchboard \citep{Jurafsky:97-damsl}, to compare uptake phenomena captured by \pit{} and \jsd{} at a large scale. We identify five uptake phenomena labeled in Switchboard and map them to SWBD-DAMSL tags: acknowledgment, answer, collaborative completion, reformulation and repetition (see details in Appendix~\ref{sec:appendix_damsl}).

We estimate scores for \pit{} and \jsd{} for all utterance pairs $(S,T)$ in Switchboard, filtering out ones where $S <$ 5 tokens. We apply our \jsd{} model from Section~\ref{ssec:nuc}, which was partially fine-tuned on Switchboard. Since both measures are bounded, we quantile-transform the distribution of each measure to have a uniform distribution. For each uptake phenomenon, we compute the difference ($\delta$) between the median score from \jsd{} and the median score from \pit{} for all $(S, T)$ pairs where $T$ is labeled for that phenomenon.

The results (Figure~\ref{fig:swbd_damsl}) show that \jsd{} predicts significantly higher scores than \pit{} for all phenomena, especially for answers, reformulations, collaborative completions and acknowledgments. For repetition, $\delta$ is quite small, but still significant due to the large sample size. These findings corroborate our hypothesis that \pit{} and \jsd{} capture repetition similarly, but \jsd{} is able to better capture other uptake phenomena.

\begin{table*}[]
\centering
\resizebox{.95\linewidth}{!}{
\begin{tabular}{@{}llllllll@{}}
\toprule
\textbf{Dataset}          & \textbf{Size}                                                   & \textbf{Genre}                                                                & \textbf{Topic}                                                           & \textbf{Class size}                                                            & \textbf{Outcome}             & \textbf{\jsd{} ($\beta$)} & \textbf{\pit{} ($\beta$)}              \\ \midrule
\multirow{2}{*}{NCTE}     & 1.6k conv.                                                      & \multirow{2}{*}{\begin{tabular}[c]{@{}l@{}}in-person\\spoken\end{tabular}} & \multirow{2}{*}{math}                                                    & \multirow{2}{*}{\begin{tabular}[c]{@{}l@{}}whole class \end{tabular}} & use of student contributions & .101***      & .113***                  \\
                          & 55k $(S, T)$                                                      &                                                                               &                                                                          &                                                                                & math instruction quality     & .091***      & .121***                  \\ \midrule
SimTeacher                & \begin{tabular}[c]{@{}l@{}}338 conv.\\ 2.7k $(S, T)$\end{tabular} & \begin{tabular}[c]{@{}l@{}}virtual\\spoken\end{tabular}                    & literature                                                               & \begin{tabular}[c]{@{}l@{}}small group\end{tabular}                      & quality of feedback          & .127*        & .123*                     \\ \midrule
\multirow{2}{*}{Tutoring} & 4.6k conv.                                                      & \multirow{2}{*}{\begin{tabular}[c]{@{}l@{}}virtual\\written\end{tabular}}  & \multirow{2}{*}{\begin{tabular}[c]{@{}l@{}}math,\\ science\end{tabular}} & \multirow{2}{*}{\begin{tabular}[c]{@{}l@{}}one-on-one\end{tabular}}      & student satisfaction         & .069***      & \multicolumn{1}{l}{.008} \\
                          & 85k $(S, T)$                                                      &                                                                               &                                                                          &                                                                                & external reviewer rating     & .063***      & \multicolumn{1}{l}{.021} \\ \bottomrule
\end{tabular}
}

\caption{The correlation of uptake scores from \jsd{} and \pit{} and outcomes for three educational datasets. The $\beta$ values represent z-scored coefficients, each obtained from an ordinary least squares regression, controlling for the number of $(S, T)$ pairs with uptake scores in each conversation (*: $p< 0.05$, **: $p<0.01$, ***: $p < 0.001$). \label{tab:outcomes}}
\end{table*}

\section{Downstream Application}
\label{sec:outcomes}

To test the generalizability of our uptake measures and their link to instruction quality, we correlate \jsd{} and \pit{} with 
educational outcomes on three different datasets of student-teacher interactions (Table~\ref{tab:outcomes}).


\paragraph{NCTE dataset.} We use all transcripts from the NCTE dataset \citep{kane2015national} (Section~\ref{sec:data}) with associated classroom observation scores based on the MQI coding instrument \citep{learning2011measuring}. We select two items from MQI relevant to uptake as outcomes: (1) use of student math contributions and (2) overall quality of math instruction. Since these items are coded at a 7-minute segment-level, we take the average ratings across raters and segments for each transcript.

\paragraph{Tutoring dataset.} We use data from an educational technology company \citep[same as in][]{chen2019predictors}, which provides on-demand text-based tutoring for math and science. With a mobile application, a student can take a picture of a problem or write it down, and is then connected to a professional tutor who
guides the student to solve the problem. Similarly to \citet{chen2019predictors}, we filter
out short sessions where the tutors are unlikely to deliver meaningful tutoring. Specifically, we create a list of $(S, T)$ pairs for all sessions, keeping pairs where $S \geq $ 5 tokens, and then remove sessions with fewer than ten $(S, T)$ pairs. This results in 4604 sessions, representing 108 tutors and 1821 students. Each session is associated with two outcome measures: (1) student satisfaction scores (1-5 scale) and (2) a rating by the tutor manager based on an evaluation rubric (0-1 scale).

\paragraph{SimTeacher dataset.} We use a dataset collected by \citet{cohen2020teacher}, via a mixed reality simulation platform in which novice teachers
get to practice key classroom skills in a virtual classroom interface populated by student avatars. The avatars are controlled remotely by a trained actor; hence the term ``mixed''
reality. All pre-service teachers from a large public university complete a five-minute simulation session at multiple timepoints in their teacher preparation program, and are coached on how to better elicit students’ thinking
about a text. We use data from Fall 2019, with 338 sessions representing 117 teachers. Since all sessions are based on the \emph{same scenario} (discussed text, leading questions, avatar scripts), this dataset uniquely allows us to answer the question: controlling for student avatar scripts, does a greater teacher uptake lead to better outcomes? For the outcome variable, we use their holistic ``quality of feedback'' measure (1-10 scale), annotated at the transcript-level by the original research team.\footnote{This overall quality
scale accounts for the extent to which teachers actively work to support student
avatars’ development of text-based responses, highlighting the importance of probing student
responses (e.g. “Where in the text did you see that?”; “What made you
think this about the character?”).}


\subsection{Results \& Analysis}

\begin{table*}[]
\centering
\resizebox{\linewidth}{!}{
\begin{tabular}{@{}ll@{}}
\toprule
\cellcolor[HTML]{9AFF99} 
\begin{tabular}[c]{@{}l@{}}\textbf{high student feedback (\pit{} $<$ \jsd{})}\end{tabular}                                                                                                                                                                                                                                                                                                                                                                                           & \cellcolor[HTML]{FFCCC9}\begin{tabular}[c]{@{}l@{}}\textbf{low student feedback (\jsd{} $<$ \pit{})}\end{tabular}                                                                                                                                                                                                                                                                                                                                                                                         \\ \midrule
\begin{tabular}[c]{@{}l@{}} S: if they're the same length i think\\ T: that's right! all we need is the length, and that's enough.\\ S: the energy from the one pendulum moving will transfer the\\ same frequency to the second pendulum once they touch?\\ T: they don't even need to touch! we can swing them so they\\ swing side by side, like two swings on a swingset.\\ S: pendulum one will start to absorb energy from pendulum two?\\ T: exactly! and eventually, the whole process will reverse until\\ pendulum one is moving full speed again.\end{tabular} & \begin{tabular}[c]{@{}l@{}}S: when you are saying mixture are you talking about nitrogen?\\ T: thanks for your question.\\ S: no i don't think so\\ T: great answer!\\ S: i don't know , just made an educated guess\\ T: great try!\\ S: i want further explanation about volume and\\ number moles when using nitrogen\\ T: sure. no worries!\end{tabular}
 \\ \bottomrule
\end{tabular}
}

\caption{Examples from the tutoring dataset --- for both examples, the predictions by \jsd{} are more accurate than the ones by \pit{} that predicts too low and too high values, respectively, when compared to student ratings. \label{tab:yup_examples}  
}
\end{table*}

As outcomes are linked to conversations, we first mean-aggregate uptake scores to the conversation-level. We then compute the correlation of uptake scores and outcomes using an ordinary least squares regression, controlling for the number of $(S, T)$ pairs in each conversation.

The results (Table~\ref{tab:outcomes}) indicate that \jsd{} correlates with all of the outcome measures significantly. \pit{} also shows significant correlations for NCTE and for SimTeacher, but not for the tutoring dataset. We provide more details below.

For NCTE and SimTeacher, we find that two measures show similar positive correlations with outcomes. These results provide further insight into our earlier findings from Section~\ref{ssec:nuc_results}. They suggest that the teacher's repetition of student words, also known as ``revoicing'' in math education \citep{forman1997learning,oconnor1993aligning}, may be an especially important mediator of instruction quality in classroom contexts and other aspects of uptake are relatively less important. The significant correlation of \jsd{} with the outcome in case of SimTeacher is especially noteworthy because \jsd{} was \emph{not} fine-tuned on this dataset (Section~\ref{ssec:nuc}); this provides evidence for the adaptability of a pre-trained model to other (similar) datasets.

The gap between the two measures in case of the tutoring dataset is an interesting finding, possibly explained by the conversational setting: repetition may be an effective uptake strategy in multi-participant \& spoken settings, ensuring that everyone has heard what the student said and is on the same page; whereas, in a written 1:1 teaching setting, repetition may not be necessary or effective as both participants are likely to assume that that their interlocutor has read their words. Our qualitative analysis suggests \jsd{} might be outperforming \pit{} because it is better able to pick up on cues related to teacher responsiveness (we include two examples in Table~\ref{tab:yup_examples}). To test this, we detect coarse-grained estimates of teacher uptake: teacher question marks (estimate of follow-up question) and teacher exclamation marks (estimate of approval). We then follow the same procedure as in Section~\ref{ssec:nuc_results} and find that dialogs where \jsd{} outperforms \pit{}, in terms of predicting student ratings, have a higher ratio of exchanges with teacher questions ($p < 0.05$, obtained from two-sample t-test) and teacher exclamation marks ($p < 0.01$).

To put these effect sizes from Table~\ref{tab:outcomes} (where significant) in the context of education interventions that are designed to increase student outcomes (typically test scores), the coefficients we report here are considered average for an effective educational intervention~\citep{kraft2020interpreting}. Further, existing guidelines for educational interventions would classify uptake as a promising potential intervention, as it is highly scalable and easily quantified.

\section{Related Work}
\label{sec:related_work}

Prior computational work on classroom discourse has employed supervised, feature-based classifiers to detect teachers' discourse moves relevant to student learning, such as authentic questions, elaborated feedback and uptake, treating these moves as binary variables \citep{samei2014domain,donnelly2017words,kelly2018automatically,stone2019utterance,jensen2020toward}. Our labeled dataset, unsupervised approach (involving a state-of-the art pre-trained model), and careful analysis across domains are novel contributions that will enable a fine-grained and domain-adaptable measure of uptake that can support researchers and teachers.


Our work aligns closely with research on the computational study of conversations. For example, measures have been developed to study constructiveness \citep{niculae2016conversational}, politeness \citep{danescu2013computational} and persuasion \citep{tan2016winning} in conversations. Perhaps most similar to our work, \citet{zhang2020balancing} develop an unsupervised method to identify therapists' backward- and forward-looking utterances, with which they guide their conversations.

We also draw on work measuring discourse coherence via embedding cosines \citep[][]{xu18,ko19},
 or via utterance classification \citep[][]{xu2019cross,iter2020pretraining}, the latter of which is used also for building and evaluating dialog systems \citep[][]{lowe2016evaluation,wolf2019transfer}.
Our work extends these two families of methods to human conversation and highlights the different linguistic phenomena they capture. Finally, our work shows the key role of coherence in the socially important task of studying uptake.
\section{Conclusion}

We propose a framework for measuring uptake, a core conversational phenomenon with particularly high relevance in teaching contexts. We release an annotated dataset and develop and compare unsupervised measures of uptake, demonstrating significant correlation with educational outcomes across three datasets. This lays the groundwork (1) for scaling up teachers' professional development on uptake thereby enabling improvements to education, (2) for conducting analyses on uptake across domains and languages where labeled data does not exist and (3) for studying the effect of uptake on a wider range of socially relevant outcomes.

\section*{Acknowledgments}
We thank anonymous reviewers, Amelia Hardy, Ashwin Paranjape, Yiwei Luo for helpful feedback. We
are grateful for the support of the Melvin and Joan Lane Stanford Graduate Fellowship (to D.D.).

\section{Ethical Considerations}

Our objective in building a dataset and a framework for measuring uptake is (1) to aid researchers studying conversations and teaching and (2) to (ultimately) support the professional development of educators by providing them with a scalable measure of a phenomenon that supports student learning. Our second objective is especially important, since existing forms of professional development aimed at improving uptake are highly resource intensive (involving classroom observations and manual evaluation). This costliness has meant that teachers working in under-resourced school systems have thus far had limited access to quality professional development in this area.

The dataset we release is sampled from transcripts collected by the National Center for Teacher Effectiveness (NCTE) \citep{kane2015national} (Harvard IRB \#17768). These transcripts represent data from 317 teachers across 4 school districts in New England that serve largely low-income, historically marginalized students. The data was collected as part of a carefully designed study on teacher effectiveness, spanning three years between 2010 and 2013 and it was de-identified by the original research team, meaning that in the transcripts, student names are replaced with ``Student'' and teacher names are replaced with ``Teacher''. Both parents and teachers gave consent for the de-identified data to be retained and used in future research. The collection process and representativeness of the data are all described in great detail in \citep{kane2015national}. Given that the dataset was collected a decade ago, there may be limitations to its use and ongoing relevance. That said, research in education reform has long attested to the fact that teaching practices have remained relatively constant over the past century \citep{cuban1993teachers,cohen2017reform} and that there are strong socio-cultural pressures that maintain this \citep{cohen1988teaching}.

The data was annotated by 13 raters, whose demographics are largely representative of teacher demographics in the US\footnote{\url{https://nces.ed.gov/fastfacts/display.asp?id=28}}. All raters have domain expertise, in that they are former or current math teachers and former or current raters for the Mathematical Quality Instruction \citep{learning2011measuring}. The raters were trained for at least an hour each on the coding instrument and spent 8 hours on average on the annotation (over the course of several weeks) and were compensated \$16.5 / hr.

In Section~\ref{sec:outcomes}, we apply our data to to two educational datasets besides NCTE. We do not release either of these datasets. The SimTeacher dataset was collected by \citet{cohen2020teacher} (University of Virginia IRB \#2918), for research and program improvement purposes. The participants in the study are mostly white (82\%), female (90\%), and middle class (71\%), mirroring the broader teaching profession. As for the tutoring dataset, the data belongs to a private company; the students and tutors have given consent for their data to be used for research, with the goal of improving the company's services. The company works with a large number of tutors and students; we use data that represents 108 tutors and 1821 students. 70\% of tutors in the data are male, complementing the other datasets where the majority of teachers are female. The company does not share other demographic information about tutors and students.

Similarly to other data-driven approaches, it is important to think carefully about the source of the training data when considering downstream use cases of our measure. Our unsupervised approach helps address this issue as it allows for training the model on data that is representative of the population that it is meant to serve.

\bibliographystyle{acl_natbib}
\bibliography{anthology,main}

\clearpage
\appendix

\section{Annotation Framework}
\label{sec:appendix_annotation}

Figure~\ref{fig:annotation_interface} shows a screenshot of our annotation interface. In the annotation framework, we used the term ``active listening'' to refer to uptake, since we found that active listening is more interpretable to raters, while uptake is too technical. However, the difference in terminology should not affect the annotations, since the two constructs are synonymous and we designed the annotation instructions entirely based on the linguistics and education literature on uptake. For example, the title of the instruction manual is ``Annotating Teachers' Uptake of Student Ideas'', and we define different levels of uptake with phrasings such as ``the teacher provides evidence for following what the student is saying or trying to say'', linking our definition to \citet{clark1989contributing}'s theory on grounding. We include annotation instructions with the dataset.

\begin{minipage}{\textwidth}
\begin{center}
\includegraphics[width=\textwidth]{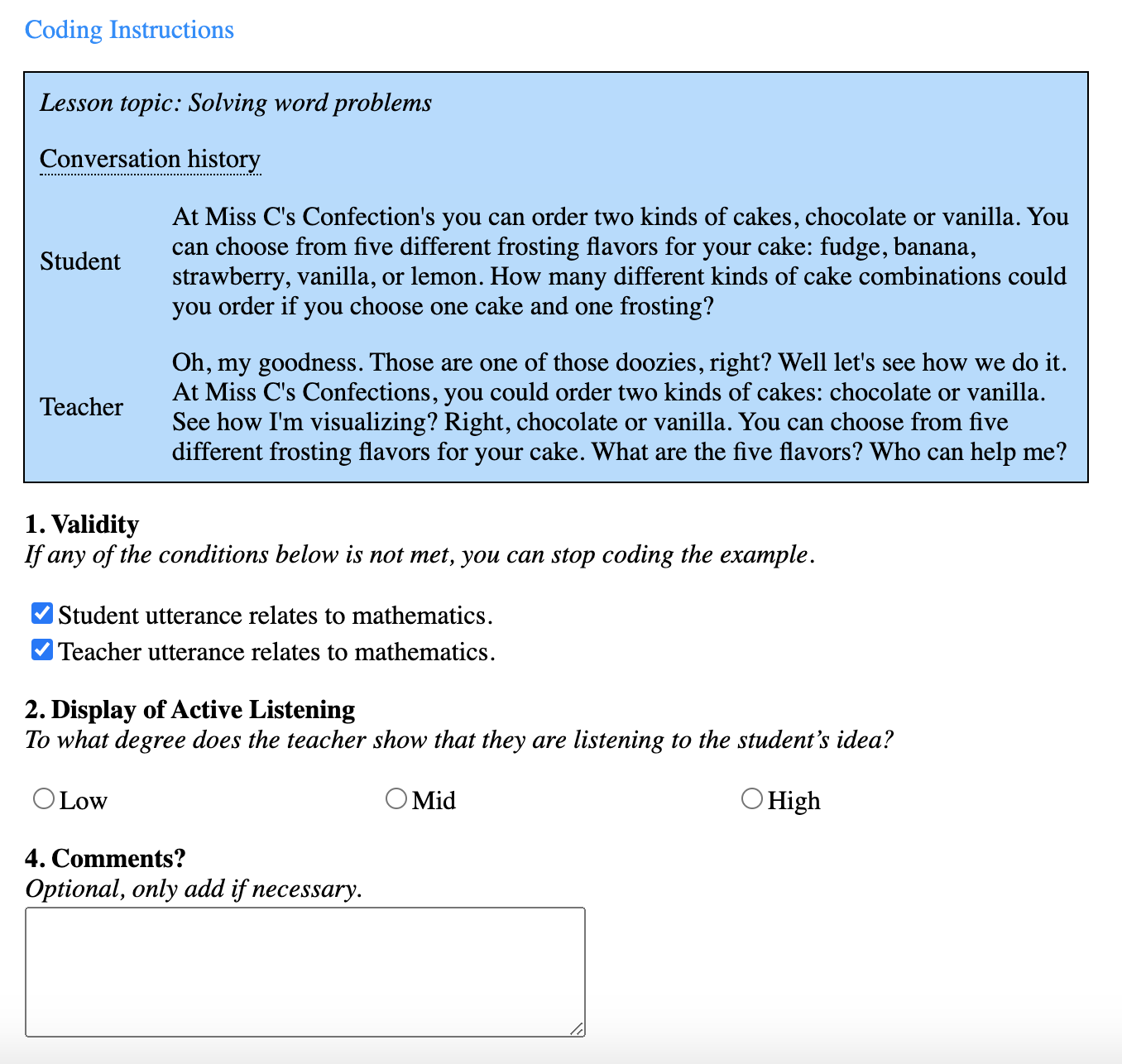}
\captionof{figure}{Screenshot of the annotation interface. \label{fig:annotation_interface}}
\end{center}
\end{minipage}

\clearpage

\section{Supervised Model Results}
\label{sec:appendix_supervised}

We conducted experiments to compare the performance of our unsupervised models to that of supervised models. We randomly split the annotated data into training (80\%) and test (20\%) sets, using the z-scored rater judgments as labels (Section~\ref{sec:data}). We trained BERT-base \citep{devlin2019bert} and RoBERTa-base \citep{liu2019roberta} on this data for 10 epochs with early stopping, and a batch size of 8 $\times$ 2 gradient accumulation steps --- all other parameters are defaults set by Huggingface\footnote{\url{https://huggingface.co/}}.

The results are shown in Table~\ref{tab:supervised_results}. The supervised models outperform our unsupervised models by less than $.08$, indicating the competitiveness of our unsupervised methods. Interestingly, we also find that BERT outperforms RoBERTa, a gap that persisted despite tuning the number of training epochs. Since our paper’s focus is unsupervised methods that enable scalability and domain-generalizability, we leave more extensive parameter search and supervised model comparison for future work.

\begin{table}[]
\centering
\begin{tabular}{@{}ll@{}}
\toprule
\textbf{Model}  & $\boldsymbol{\rho}$ \\ \midrule
\jsd &   $.540$    \\ \midrule
RoBERTa-base &   $.561$          \\ 
BERT-base    &  $.618$    \\
\bottomrule

\end{tabular}
\caption{Supervised model results. \label{tab:supervised_results}}
\end{table}

\section{Mapping the SWBD-DAMSL Tagset to Uptake Phenomena}
\label{sec:appendix_damsl}

We map tags from SWBD-DAMSL~\citep{Jurafsky:97-damsl} to five salient uptake phenomena: acknowledgment, answer, reformulation, collaborative completion and repetition. Table~\ref{tab:damsl_mapping} summarizes our mapping. Since acknowledgment is highly frequent and it can co-occur with several other dialog acts, we consider those examples to be acknowledgments that are labeled exclusively for this phenomenon (using either the tag \emph{b}, \emph{bh} or \emph{bk}).

\begin{table*}[t!]
\centering
\begin{tabular}{@{}llr@{}}
\toprule
\textbf{Uptake phenomenon} & \textbf{DAMSL Tags} & \multicolumn{1}{l}{\textbf{\% of Examples}} \\ \midrule
acknowledgment             & b, bh, bk           & 81\%                                        \\ \midrule
answer                     & tags containing ``n'' & 13\%                                        \\ \midrule
reformulation              & bf                  & 2\%                                         \\ \midrule
collaborative completion   & \textasciicircum{}2 & 2\%                                         \\ \midrule
repetition                 & \textasciicircum{}m & 2\%                                         \\ \bottomrule
\end{tabular}
\caption{Mapping between uptake phenomena and tags from SWBD-DAMSL \citep{Jurafsky:97-damsl}. \label{tab:damsl_mapping}}
\end{table*}

\end{document}